\documentclass[letterpaper]{article} % DO NOT CHANGE THIS
\usepackage{aaai2026}  % DO NOT CHANGE THIS
\usepackage{times}  % DO NOT CHANGE THIS
\usepackage{helvet}  % DO NOT CHANGE THIS
\usepackage{courier}  % DO NOT CHANGE THIS
\usepackage{graphicx} % DO NOT CHANGE THIS
\usepackage{natbib}  % DO NOT CHANGE THIS AND DO NOT ADD ANY OPTIONS TO IT
\usepackage{caption} % DO NOT CHANGE THIS AND DO NOT ADD ANY OPTIONS TO IT
\usepackage{algorithm}
\usepackage{algorithmic}
\usepackage[dvipsnames]{xcolor}
\usepackage{amsmath}
\usepackage{amssymb}
\usepackage{booktabs} 
\usepackage{subfig}
\usepackage{multirow}
\usepackage{rotating}
\usepackage{adjustbox}
\usepackage[pagebackref=true,breaklinks=true,colorlinks=true,citecolor=NavyBlue,bookmarks=false]{hyperref}

\newcommand\eg{\textit{e.g.}}
\newcommand\ie{\textit{i.e.}}

\usepackage{newfloat}
\usepackage{listings}
\DeclareCaptionStyle{ruled}{labelfont=normalfont,labelsep=colon,strut=off} % DO NOT CHANGE THIS
\floatstyle{ruled}
\newfloat{listing}{tb}{lst}{}
\floatname{listing}{Listing}
\title{Steering One-Step Diffusion Model with Fidelity-Rich Decoder \\for Fast Image Compression}

\author{
    Zheng Chen$^{1}$\equalcontrib,\enspace 
    Mingde Zhou$^{1}$\equalcontrib,\enspace
    Jinpei Guo$^{2}$,\enspace\\
    Jiale Yuan$^{1}$,\enspace
    Yifei Ji$^{1}$,\enspace
    Yulun Zhang$^{1}$\thanks{Corresponding author: Yulun Zhang, yulun100@gmail.com}
}
\affiliations{
  \textsuperscript{1}Shanghai Jiao Tong University,\enspace
  \textsuperscript{2}Carnegie Mellon University
}

\usepackage{bibentry}
\begin{document}

\maketitle

\begin{abstract}
Diffusion-based image compression has demonstrated impressive perceptual performance. However, it suffers from two critical drawbacks: \textbf{(1)} excessive decoding latency due to multi-step sampling, and \textbf{(2)} poor fidelity resulting from over-reliance on generative priors. To address these issues, we propose SODEC, a novel single-step diffusion image compression model. We argue that in image compression, a sufficiently informative latent renders multi-step refinement unnecessary. Based on this insight, we leverage a pre-trained VAE-based model to produce latents with rich information, and replace the iterative denoising process with a single-step decoding. Meanwhile, to improve fidelity, we introduce the fidelity guidance module, encouraging output that is faithful to the original image. Furthermore, we design the rate annealing training strategy to enable effective training under extremely low bitrates. Extensive experiments show that SODEC significantly outperforms existing methods, achieving superior rate-distortion-perception performance. Moreover, compared to previous diffusion-based compression models, SODEC improves decoding speed by more than 20$\times$. Code is released at:~\url{https://github.com/zhengchen1999/SODEC}.
\end{abstract}


%     \link{Code}{https://github.com/zhengchen1999/SODEC}
%     \link{Extended version}{https://arxiv.org/abs/2508.04979}

\vspace{-4.mm}
\section{Introduction}
The rising cost of data storage and transmission underscores the importance of image compression. Traditional codecs such as JPEG2000~\cite{taubman2002jpeg2000} and VVC~\cite{bross2021overview} perform reliably at medium to high bitrates. However, when the bitrate drops to low levels (\eg, $<$0.1 bpp), they tend to produce block artifacts, blurring, and structural distortions. Achieving a balance between distortion and perceptual quality under low bitrate constraints remains a challenging problem.

\begin{figure}[t]
\centering
\includegraphics[width=\linewidth]{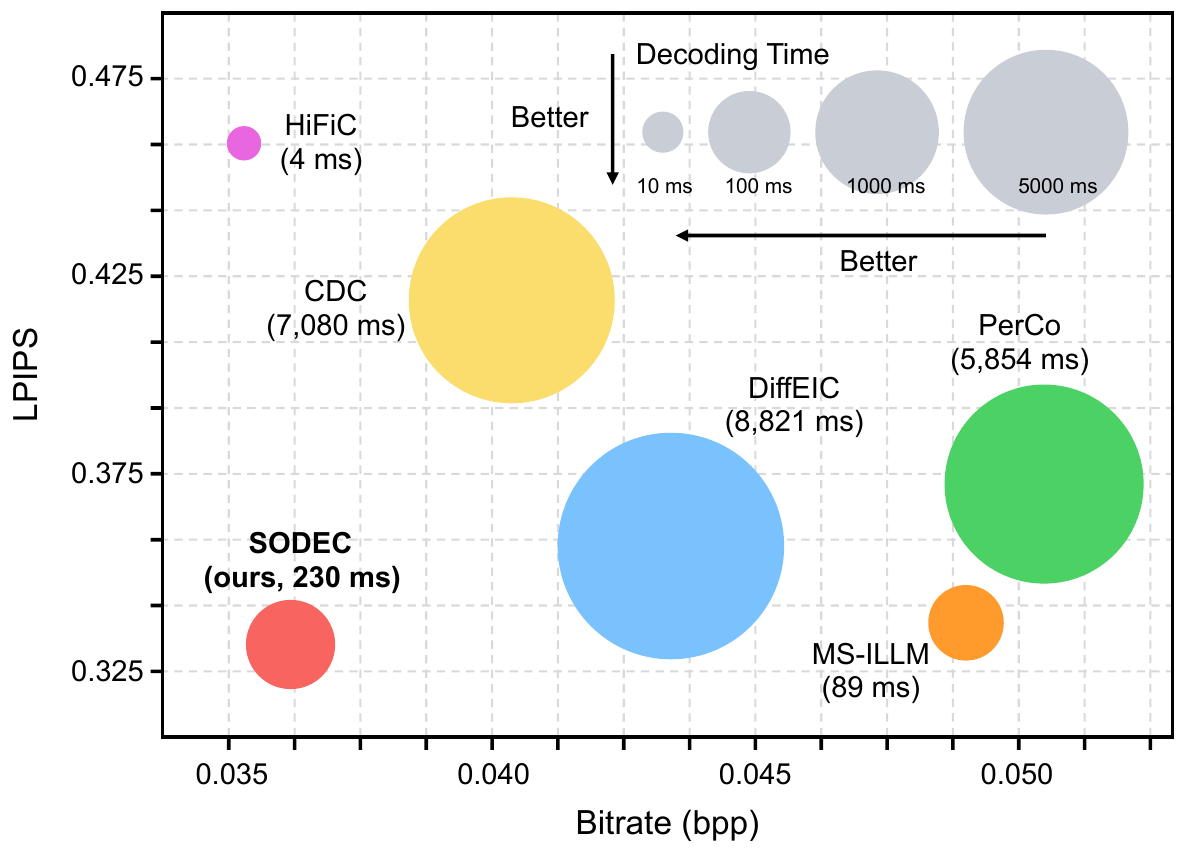}
\vspace{-7.mm}
\caption{
LPIPS-bitrate-latency comparison on DIV2K-Val. Decoding time is measured on 512$\times$512 images using one A6000 GPU. Our method achieves the best perceptual quality (\ie, LPIPS). Meanwhile, compared to the multi-step diffusion-based method DiffEIC~\cite{li2024towards}, our method offers a 38$\times$ speedup in decoding time.}
\label{fig:performance}
\vspace{-6.mm}
\end{figure}

In recent years, learning-based image compression models built upon variational autoencoders (VAEs)~\cite{kingma2019introduction} have surpassed traditional methods in the rate-distortion trade-off~\cite{balle2018variational,cheng2020learned,van2017neural}. Benefiting from advances in probabilistic modeling, such as hyperpriors~\cite{balle2018variational,minnen2018joint}, these approaches typically excel in distortion-oriented metrics like PSNR and MS-SSIM. Moreover, to better align with human perception, subsequent works further incorporate perception-oriented objectives, leading to a more comprehensive rate-distortion-perception framework~\cite{blau2019rethinking,mentzer2020high,muckley2023improving,agustsson2023multi,he2022po}. These methods achieve a more realistic reconstruction by employing distortion and perceptual losses to enhance realism. However, VAE-based methods struggle to reconstruct details when operating at extremely low bitrates, resulting in poor perceptual quality. In other words, while the reconstructions may appear ``technically correct'', they often lack realism.

In contrast, diffusion models~\cite{ho2020denoising} have recently demonstrated remarkable performance in the rate-perception trade-off, due to their powerful generative priors. Specifically, in diffusion-based methods, the encoder produces a compact latent representation, while decoding is reformulated as a multi-step conditional denoising process~\cite{theis2022lossy,lei2023textsketch}. Guided by conditional signals derived from the bitstream, the diffusion model iteratively refines a noisy latent~\cite{yang2023lossy,vonderfecht2025lossy,careil2024towards,ghouse2023residual,relic2024lossy}. Thus, diffusion-based models can synthesize highly realistic textures and details, even under extreme compression. Moreover, some approaches integrate global (\eg, text prompts) or local (\eg, quantized features) guidance to constrain the generative process~\cite{pan2022extreme,careil2023towards,li2024towards}. 
% These conditions guide the model to generate more plausible details, yielding more realistic reconstructions.

However, such models face two critical challenges:
\textbf{(1) High latency.} The multi-step denoising process incurs substantial decoding latency and computational cost. This limits their applicability in real-time or resource-constrained scenarios.
\textbf{(2) Low fidelity.} The generative nature of diffusion models makes them heavily reliant on pre-trained priors rather than the input itself. This leads to reconstructions that deviate from the original content, compromising fidelity.

To address these challenges, we propose SODEC (steering one-step diffusion model with fidelity-rich decoder), a novel image compression model designed for low-bitrate scenarios. Our SODEC is designed around efficient decoding and high-fidelity guidance.
\textbf{(1) Single-step decoding.} To mitigate the high latency of multi-step diffusion, we replace the iterative denoising process with a single-step process. Benefits to the informative latent representations produced by the pre-trained VAE-based compression model, single-step decoding is sufficient to realize high-quality reconstruction.
\textbf{(2) Fidelity guidance module.} To compensate for the potential fidelity loss, we employ a pre-trained VAE-based compression model to produce a high-fidelity preliminary reconstruction. This reconstruction serves as explicit visual guidance to the diffusion model, encouraging outputs faithful to the source image.
\textbf{(3) Rate annealing training strategy.} To ensure effective training at extremely low bitrates, we adopt a three-stage optimization. The model is first pre-trained at higher bitrates to learn informative representations. Then, we gradually anneal the model to the target bitrate, selectively preserving essential information.

Benefits to above designs, SODEC achieves impressive performance in terms of rate-distortion-perception trade-off. Furthermore, due to the single-step and lightweight conditioning, our SODEC achieves excellent decoding efficiency. As shown in Fig.~\ref{fig:performance}, compared to multi-step diffusion paradigms (\eg, PerCo~\cite{careil2023towards}, DiffEIC~\cite{li2024towards}), SODEC delivers over 20$\times$ speedup.

Our contributions are summarized as follows:

\begin{itemize}
\item We propose SODEC, a single-step diffusion image compression model that significantly accelerates decoding while preserving high perceptual-fidelity quality.

\item We introduce the fidelity guidance module, a diffusion guidance mechanism conditioned on high-fidelity reconstruction, effectively improving content fidelity.

\item We develop the rate annealing training strategy, a three-stage optimization scheme that enables the model to retain critical information at extremely low bitrates.

\item SODEC achieves state-of-the-art performance in the rate-distortion-perception trade-off, while delivering significantly improved decoding efficiency.
\end{itemize}

\section{Related Work}
\subsection{VAE-based Compression Model}
Compressing images at extremely low bitrates is a challenge where traditional methods like JPEG2000~\cite{taubman2002jpeg2000} and VVC~\cite{bross2021overview} often produce severe blurring and artifacts. Recently, learned compression based on Variational Autoencoders (VAEs)~\cite{kingma2013auto} has surpassed traditional codecs in rate-distortion performance~\cite{balle2018variational,cheng2020learned,wang2022neural,minnen2020channel,he2022elic}, largely due to innovations like the hyperprior model. This architecture is refined with sophisticated context models and quantization strategies, such as the hierarchical prior model~\cite{minnen2018joint} and VQ-VAE~\cite{van2017neural}, achieving state-of-the-art performance on distortion-oriented metrics like PSNR and MS-SSIM. Subsequently, to enhance visual realism, perception-oriented models~\cite{tschannen2018deep,blau2019rethinking,agustsson2019generative,mentzer2020high,muckley2023improving} are introduced to optimize the rate-distortion-perception. However, these models still tend to produce artifacts and lack detail at extremely low bitrates.

\subsection{Diffusion-based Compression Model}
Diffusion models~\cite{ho2020denoising,song2020denoising,rombach2022high} excel at high-quality image synthesis by framing generation as an efficient, latent-space noise prediction task. Recent compression works adapt these models for image compression by treating it as a conditional denoising problem~\cite{saharia2021image,xia2025diffpc,liu2024extreme}. Typically, an encoder transforms the source image into a compact latent representation that conditions the reverse diffusion process, enabling reconstructions with high perceptual quality. This paradigm is demonstrated by foundational methods like CDC~\cite{yang2023lossy}, which conditions on a learned latent. More sophisticated strategies, \eg, DiffC~\cite{vonderfecht2025lossy}, use reverse-channel coding to steer a pre-trained diffusion model without fine-tuning. 

Moreover, diffusion-based compression models employ various guidance signals to enhance reconstruction quality. For example, some approaches compress images into a purely semantic space~\cite{lei2023text,bachard2024coclico,pan2022extreme}. For instance, Pan et al.~\cite{pan2022extreme} encode an image into a textual embedding that subsequently guides a pre-trained text-to-image model. Other works~\cite{careil2023towards,guo2025oscar,li2024towards} utilize more sophisticated conditioning. For example, PerCo applies both pre-extracted text prompts for global context and quantized visual features for local details. In contrast, DiffEIC derives its guidance internally, extracting a global context vector from the hyperprior and injecting it into the diffusion process.

However, these methods share two primary limitations: \textbf{(1)} the substantial latency from their multi-step diffusion process, and \textbf{(2)} the tendency to sacrifice fidelity for perceptual realism due to the diffusion prior.

\begin{figure*}[t]
\centering
\includegraphics[width=\textwidth]{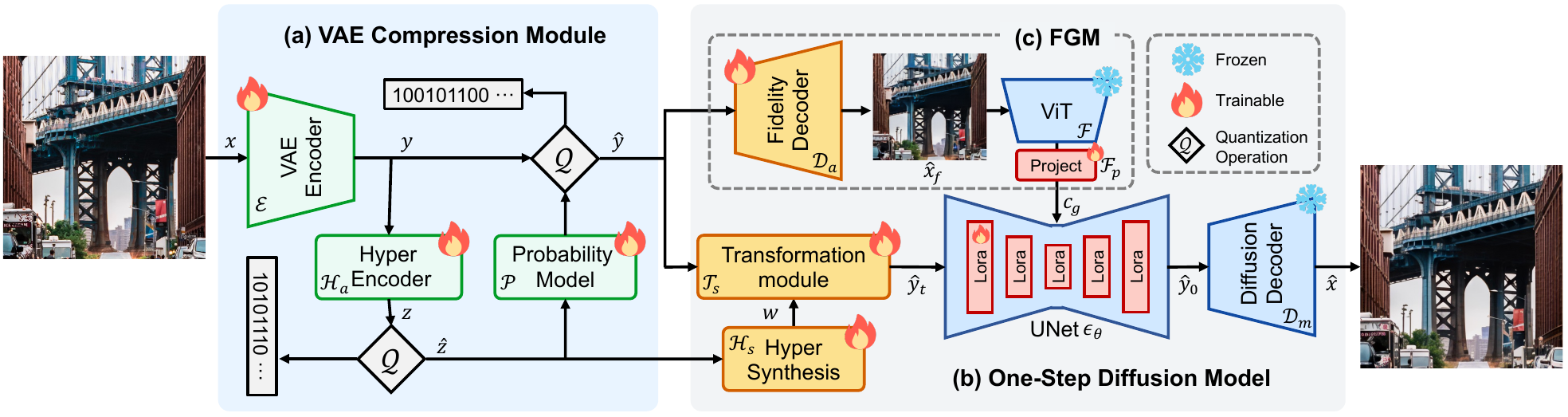}
\caption{Overview of SODEC. (a) VAE compression module: A pre-trained VAE-based compression model is used to generate the informative latent representation. (b) One-step diffusion model: The latent is mapped to the diffusion space via the transformation module, followed by single-step denoising to produce the reconstructed output. (c) Fidelity guidance module (FGM): A high-fidelity preliminary reconstruction is generated using the VAE-based compression model. Then, the pre-trained ViT is used to extract visual features as the guidance for the diffusion model.}
\vspace{-3.mm}
\label{overview}
\end{figure*}

\section{Methodology}
In this section, we provide an overview of our proposed model, SODEC, as illustrated in Fig.~\ref{overview}.
The section begins with the VAE Compression Module. Subsequently, we elaborate on the core component of SODEC: the one-step diffusion model and the fidelity guidance module. Finally, we detail our rate annealing training strategy.

\subsection{SODEC Overview}
The overview of SODEC is illustrated in Fig.~\ref{overview}. The framework begins with a VAE Compression Module that downsamples a raw image $x \in \mathcal{R}^{(H\times W\times 3)}$ for 16 times to a compact latent representation $y \in \mathcal{R}^{(H/16 \times W/16 \times C)}$, where C is the latent channels (usually 220). 
After the entropy coding, restored $\hat{y}$ and $\hat{z}$ are passed into a transformation module $\mathcal{T}_s$ and converted into a content variable $\hat{y}_t \in \mathcal{R}^{(64 \times 64 \times 4)}$ suitable for diffusion process. Then, we apply the one-step diffusion model to speed up the denoising time compared to the previous multi-step diffusion model~\cite{careil2023towards,li2024towards}. The one-step diffusion model will then be used to generate the denoised content variable $\hat{y}_0$. 

Simultaneously, we utilize a pre-trained fidelity-rich decoder $\mathcal{D}_a$ and further fine-tune it for high fidelity. To achieve this goal, we introduce an alignment loss $\mathcal{L}_{align}$ that consists of pixel-wise loss that constrains $\mathcal{D}_a$ to consistently produce high-fidelity images.
After $\mathcal{D}_a$ decodes the latent representation $\hat{y}$ into the raw image $\hat{x}_f$, we use the pre-trained ViT model~\cite{liu2021swin} to capture the high-fidelity feature information. Then we linearly project it into the embedding space, getting the condition guidance $c_g$.

To achieve the best performance, we also introduce the rate annealing training strategy. This strategy first pretrains a complete VAE model with a higher bitrate than our final aim. This VAE model comprises a rich representation in the latent space. Then, we lift the rate penalty by applying a larger trade-off parameter $\lambda$ in the loss function. Thus, the model can ``distill" from the rich representation and selectively discard non-essential information. This strategy is proven to achieve better performance than directly training.

\subsection{VAE Compression Module}
The proposed SODEC employs a VAE-based compression backbone to efficiently encode the input image into a bitstream. This module is comprised of the encoder $\mathcal{E}$, hyperencoder $\mathcal{H}_a$, and probability model $\mathcal{P}$.

Given an input image $x$, the encoder $\mathcal{E}$ produces a compact latent representation $y$$=$$\mathcal{E}(x)$. Hyperencoder $\mathcal{H}_a$ then extracts the hyper-latent $z$$=$$\mathcal{H}_a(y)$. Next, both of these representations $y$ and $z$ are quantized into $\hat{y}$$=$$\mathcal{Q}(y)$, $\hat{z}$$=$$\mathcal{Q}(z)$, where $\mathcal{Q}(\cdot)$ represents the quantization operation.
Finally, the learned probability model $\mathcal{P}$ conditions on the quantized hyperprior $\hat{z}$ to predict the parameters $(\mu, \sigma)$ of a Gaussian distribution, which models the probability of latent representation $\hat{y}$ for efficient entropy coding.

For the compression model, we pre-train HiFiC~\cite{mentzer2020high} and use its learned weights to initialize our compression backbone $\mathcal{E}$, $\mathcal{H}_a$, and $\mathcal{P}$. In addition, we use the pre-trained VAE decoder to initialize the decoder $\mathcal{D}_a$ in the fidelity guidance module. We apply $\mathcal{D}_a$ to generate the high-fidelity preliminary reconstruction $\hat{x}_f$: 
\begin{equation}
\label{afdggfjhhdsgffdgjfkhglkjh}
\hat{x}_f = \mathcal{D}_a(\mathcal{Q}(\mathcal{E}(x))).
\end{equation}
This model is optimized using the rate-distortion function:
\begin{equation}
\label{19uehu91uheuqbd10101}
\mathcal{L}_{EG} = \mathbb{E}_{x \sim p_x} \left[ \lambda \cdot r(\hat{y},\hat{z}) + d(x, \hat{x}_f) \right],
\end{equation}
where  $r(\cdot)$ denotes the rate and $\lambda$ is the hyperparameter to control rate penalty, and $d(x, \hat{x}_f)$ represents the distortion:
\begin{equation}
\label{123u89018ueuh81uh8dbu1uhe}
d(x, \hat{x}_f) = k_M \cdot \mathrm{MSE}(x, \hat{x}_f) + k_P \cdot d_P(x, \hat{x}_f),
\end{equation}
where $k_M$ and $k_P$ are hyperparameters. We choose LPIPS for the ``perception distortion" $d_p$ (in all the subsequent training, we also choose LPIPS by default).

It is worth noting that, in this pre-training stage, we adopt a smaller $\lambda$ (\ie smaller rate penalty) to train a stronger VAE encoder-decoder pair with higher bitrates. This is beneficial for our subsequent training. More details will be shown in the ``Rate Annealing Training Strategy" section.

\begin{figure}[t]
\centering
\includegraphics[width=\linewidth]{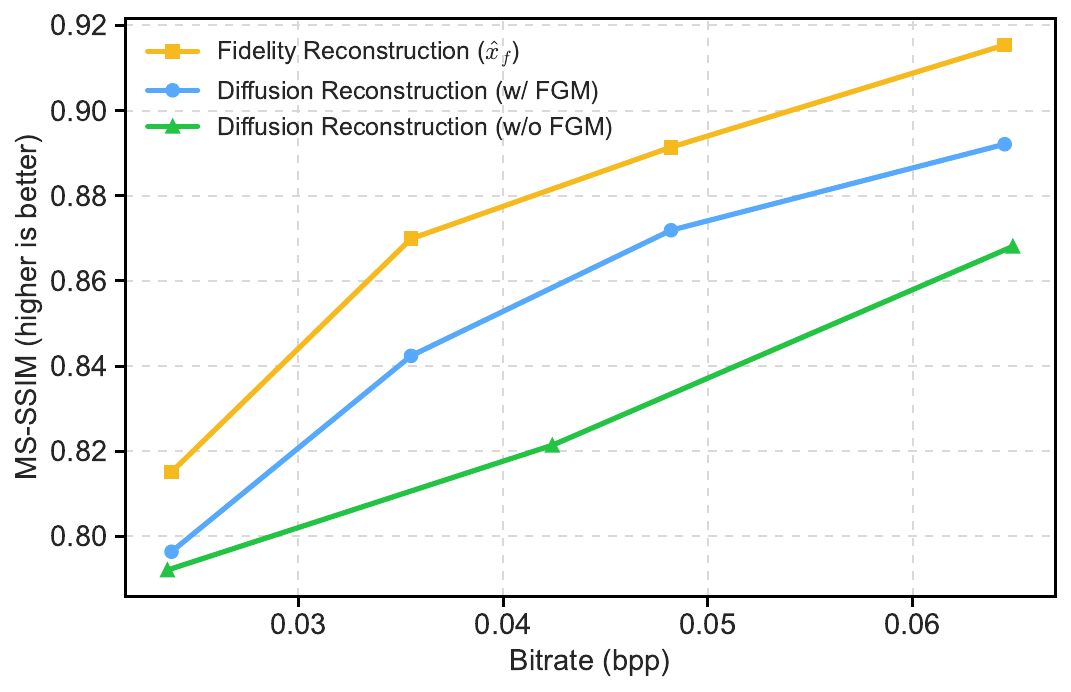}
\vspace{-5.mm}
\caption{Fidelity comparison (\ie, MS-SSIM) on DIV2K-Val. We compare MS-SSIM (with GT) under different bitrates for the fidelity reconstruction and the diffusion outputs with (w/) and without (w/o) the fidelity guidance module (FGM). The use of FGM improves reconstruction fidelity.}
\vspace{-3.mm}
\label{fidelity comparsion}
\end{figure}

\subsection{One-Step Diffusion Model}
Given $\hat{y}$ and $\hat{z}$ from the bitstream, we propose a transformation module to convert them into a content variable $\hat{y}_t$, which is suitable for diffusion denoising. 

First, we use a hyper synthesis network $\mathcal{H}_s$ to extract global information $w$ from the hyperprior $\hat{z}$, where $w$$=$$\mathcal{H}_s(\hat{z})$. Then, we then merge $w$ and $\hat{y}$ and convert them into content variables $\hat{y}_t$$=$$\mathcal{T}_s(\hat{y}, w)$, where $\mathcal{T}_s$ denotes the transformation module. Here, $\hat{y}_t$ is conceptually analogous to a noisy latent at the timestep $t$ from the forward process:
\begin{equation}
\hat{y}_t = \sqrt{\bar{\alpha}_t} \hat{y}_0 + \sqrt{1 - \bar{\alpha}_t} \epsilon.
\end{equation}
For standard diffusion models, they perform a multi-step diffusion process to predict a clear version of a noisy latent.
However, these processes are extremely slow and are the most time-consuming steps during the image reconstruction process. Thus, to speed up the diffusion process, we introduce the one-step diffusion model, based on Stable Diffusion 2.1~\cite{rombach2022high}. In this diffusion model, a noise estimator with a UNet architecture $\epsilon_{\theta}$ is used to predict the clear, denoised version of the content variable $\hat{y}_{0}$:
\begin{equation}
\label{abjijdbui1ibdbudbusjadja}
\hat{y}_0=\frac{\hat{y}_t-\sqrt{1-\bar{\alpha}_t} \epsilon_{\theta}(\hat{y}_t, t, c_g)}{\sqrt{\bar{\alpha}_t}},
\end{equation}
where $c_g$ is the condition guidance. We describe the details of $c_g$ in the next section. Finally, a pre-trained diffusion decoder $\mathcal{D}_m$ reconstructs the output image $\hat{x}$ from the denoised content variable $\hat{y}_0$, where $\hat{x}$$=$$\mathcal{D}_m(\hat{y}_0)$.
In SODEC, we set the timestep $t$ as 999. Meanwhile, to adapt the diffusion model to image compression tasks, we adopt LoRA~\cite{hu2022lora} to fine-tune the diffusion model.

\subsection{Fidelity Guidance Module}
The powerful generative prior of diffusion models enables the synthesis of high-perceptual-quality images. However, it often comes at the cost of reconstruction fidelity. To address this limitation, we propose the fidelity guidance module that injects high-fidelity information into the diffusion process. 

As shown in Fig.~\ref{fidelity comparsion}, the preliminary reconstruction $\hat{x}_f$ is highly faithful to the original image, although it may lack perceptual richness. Conversely, while the diffusion model excels at synthesizing realistic textures, it lacks explicit knowledge of the source image. Therefore, we can apply the high-fidelity reconstruction $\hat{x}_f$ as a strong conditional guide, to steer the diffusion generator to reconstruct details that are plausible and consistent with the original content. Thus, we can achieve both good fidelity and perception results.

Specifically, the module first utilizes a pre-trained fidelity-rich decoder, $\mathcal{D}_a$, to generate the high-fidelity preliminary reconstruction $\hat{x}_{f}$ from the compressed latent $\hat{y}$:
\begin{equation}
    \hat{x}_{f} = \mathcal{D}_a(\hat{y}),
    \label{eq:preliminary_recon}
\end{equation}
where $\mathcal{D}_a$ comes from the pre-trained HiFiC encoder-decoder pair, as shown in Eq.~\eqref{19uehu91uheuqbd10101}. 

Subsequently, a pre-trained ViT Transformer~\cite{dosovitskiy2020image}, denoted as the feature extractor $\mathcal{F}$, is employed to capture deep visual features from this intermediate image. These features are then mapped into the conditioning space of the diffusion model by a projection network $\mathcal{F}_{p}$ to produce the final guidance condition $c_g$:
\begin{equation}
    c_g = \mathcal{F}_{p}(\mathcal{F}(\hat{x}_{f})),
    \label{eq:guidance_generation}
\end{equation}
where the resulting condition $c_g$$\in$$\mathcal{R}^{L \times D}$ consists of a sequence of $L$ embedding vectors of dimension $D$. In our model, L and D are chosen as 77 and 1024.

This high-fidelity guidance $c_g$, which encapsulates rich high-fidelity structural information from the source, is then injected into the diffusion denoising model $\epsilon_{\theta}$ through cross-attention to steer the diffusion process.

Table~\ref{tab:ablation_guidance} in the ablation study demonstrates that this guidance mechanism can effectively steer the generative process, ensuring the final output is both perceptually realistic and highly faithful to the original content.

\subsection{Rate Annealing Training Strategy}
\label{sec: training strategy}
We propose a three-stage training strategy for our SODEC, illustrated in Fig.~\ref{overview}. This idea is based on the motivation that selecting from a rich representation and discarding non-essential information is easier than recreating detailed information. Thus, we decide to first train a VAE model with higher bitrates and then increase the rate penalty to force the model to discard and choose the most useful information.

\paragraph{Stage 1: High-Bitrate VAE Pre-training.}
Our strategy begins by pre-training HiFiC~\cite{mentzer2020high} model, which serves as the core compression component. In this stage, the model is trained end-to-end on the rate-distortion function as shown in Eq.~\eqref{19uehu91uheuqbd10101}, \ie, $\mathcal{L}_{EG} = \mathbb{E}_{x \sim p_x} \left[ \lambda \cdot r(y) + d(x, \hat{x}_f) \right]$. We intentionally use a small value for the Lagrange multiplier $\lambda$ to place a lower penalty on the bitrate. This encourages the model to learn a rich and comprehensive latent representation by prioritizing high-fidelity reconstructions. This pre-training phase is conducted on the HiFiC. After this stage, we obtain the high-bitrate version of networks $\mathcal{E}$, $\mathcal{H}_a$, $\mathcal{P}$, and $\mathcal{D}_a$.

\paragraph{Stage 2: Diffusion Path Warm-up.}
In the second stage, we transfer the learned weights of the VAE components ($\mathcal{E}, \mathcal{H}_a, \mathcal{P}, \mathcal{D}_a$) into our SODEC architecture, as shown in Fig.~\ref{overview}. The entire VAE encoding module ($\mathcal{E}, \mathcal{H}_a, \mathcal{P}$)  is frozen. The gradient flow is shown as follows:
\begin{equation}
\label{joabsdniajidjaskjndknasd}
\begin{array}{c}
\hat{x}_f = \mathcal{D}_a(\text{sg}(\mathcal{Q}(\mathcal{E}(x)))), \\[5pt]
w = \mathcal{H}_s(\text{sg}(\hat{z})),
\end{array}
\end{equation}
where ``$\text{sg}$'' denotes the stop-gradient operation, which cuts off the backpropagation of the gradient for this path.

Training is focused exclusively on the diffusion-based generator and path. Specifically, we freeze the well pre-trained model ViT and diffusion decoder $\mathcal{D}_m$ and fine-tune the UNet in diffusion using LoRA. Moreover, we train the following networks with full parameter updating: hyper synthesis network $\mathcal{H}_s$, transformation module $\mathcal{T}_s$, fidelity guidance decoder $\mathcal{D}_a$, and linear projection network $\mathcal{F}_{p}$.
The optimization objective for this stage only includes a distortion loss between the output $\hat{x}$ and the original image $x$:
\begin{equation}
\label{aokdjiqwdanddjsncjohijqkdjnak}
\mathcal{L} = \mathbb{E}_{x \sim p_x} \left[  d(x, \hat{x}) \right],
\end{equation}
where $d(\cdot)$ is the same as Eq.~\eqref{123u89018ueuh81uh8dbu1uhe}. Particularly, we do not apply a rate penalty nor an alignment loss $\mathcal{L}_{align}$, because the VAE module is frozen and the latent representation $\hat{y}$ is not distorted.
This training stage aims to teach the one-step diffusion generator to effectively map the fixed latent representations to high-quality reconstructions.

\begin{figure*}[t]
\centering
\begin{tabular}{@{}c@{\hspace{2mm}}c@{}}
    \rotatebox{90}{\ \ \ \ \ \ \ \ \ \ \ \ \ \ \ \ Kodak} & 
    \includegraphics[width=0.97\textwidth]{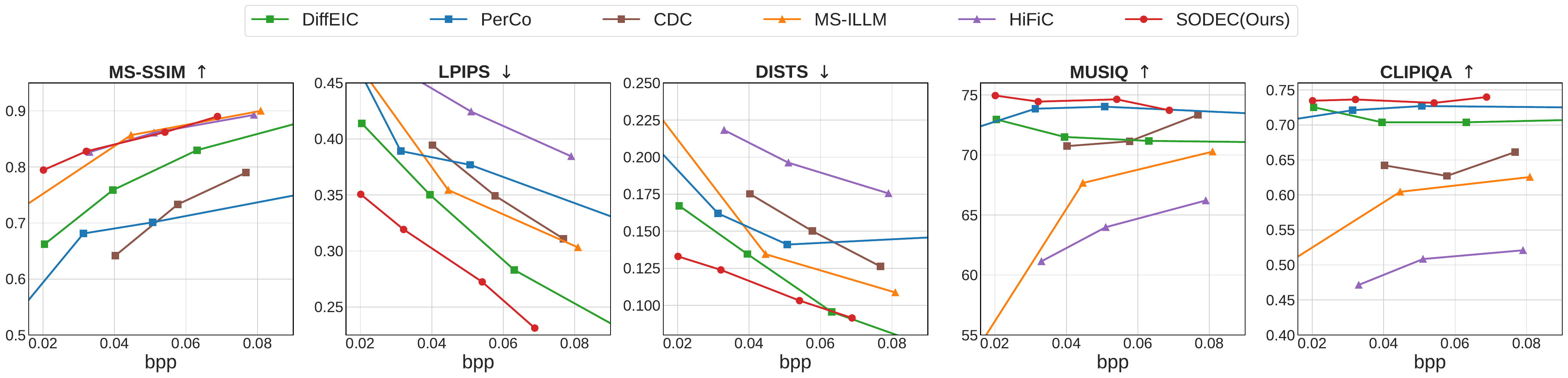} \\
    \addlinespace[5pt]
    
    \rotatebox{90}{\ \ \ \ \ \ \ \ \ \ \ DIV2K-Val} & 
    \includegraphics[width=0.97\textwidth]{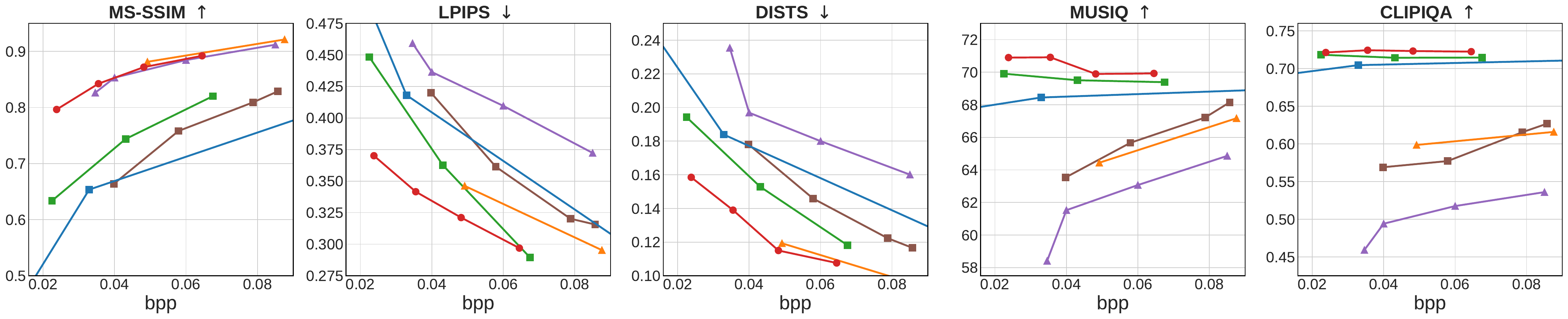} \\
    \addlinespace[5pt]
    
    \rotatebox{90}{\ \ \ \ \ \ \ \ \ \ \ \ CLIC2020} & 
    \includegraphics[width=0.97\textwidth]{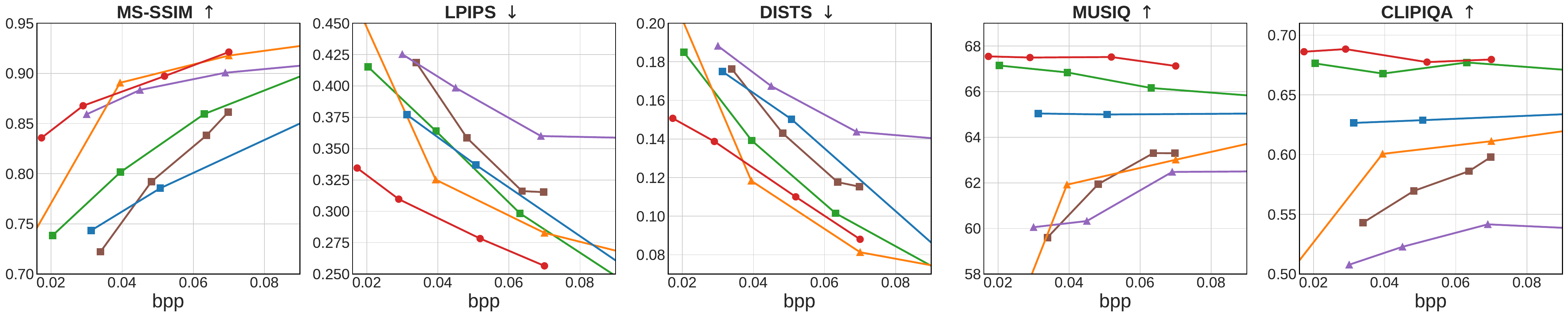}

\end{tabular}
\vspace{-2.mm}
\caption{Quantitative comparison with state-of-the-art methods on the Kodak, DIV2K-Val, and CLIC2020 datasets.}
\label{fig:main_result}
\end{figure*}

\begin{figure*}[t]
\centering
\scriptsize 
\setlength{\tabcolsep}{0.5mm} 
\renewcommand{\arraystretch}{1.4}
\begin{tabular}{ccccccccccc}

    Dataset & Original & HiFiC & CDC & MS-ILLM & PerCo & DiffEIC & \textbf{SODEC (ours)} \\
    \vspace{-2.mm}

    \includegraphics[width=0.121\textwidth]{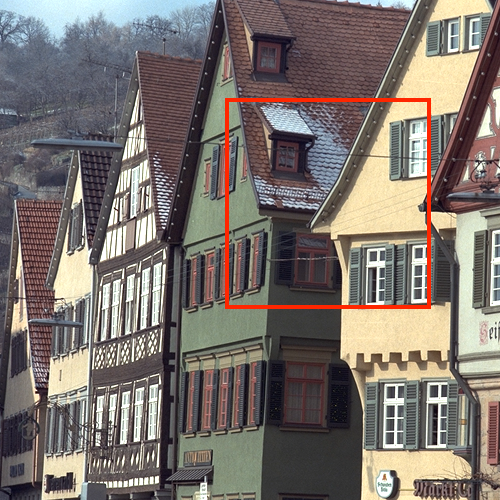} \hspace{-1.mm} &
    \includegraphics[width=0.121\textwidth]{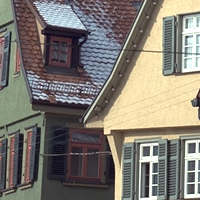} \hspace{-1.mm} &
    \includegraphics[width=0.121\textwidth]{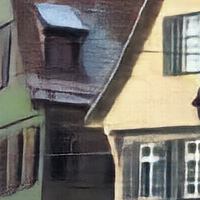} \hspace{-1.mm} &
    \includegraphics[width=0.121\textwidth]{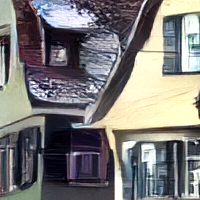} \hspace{-1.mm} &
    \includegraphics[width=0.121\textwidth]{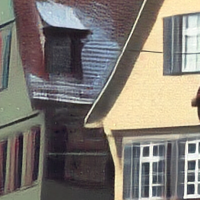} \hspace{-1.mm} &
    \includegraphics[width=0.121\textwidth]{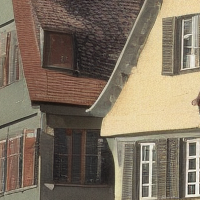} \hspace{-1.mm} &
    \includegraphics[width=0.121\textwidth]{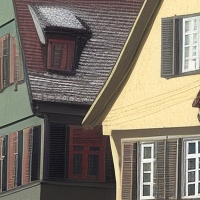} \hspace{-1.mm} &
    \includegraphics[width=0.121\textwidth]{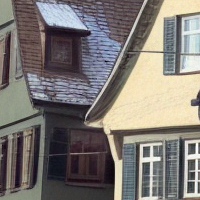} \\
    Kodak & bpp $\downarrow$ & 0.0461 & 0.0513 & 0.0399 & 0.0512 & 0.0402 & \textbf{0.0364} \\
    \vspace{-2.mm}
    
    \includegraphics[width=0.121\textwidth]{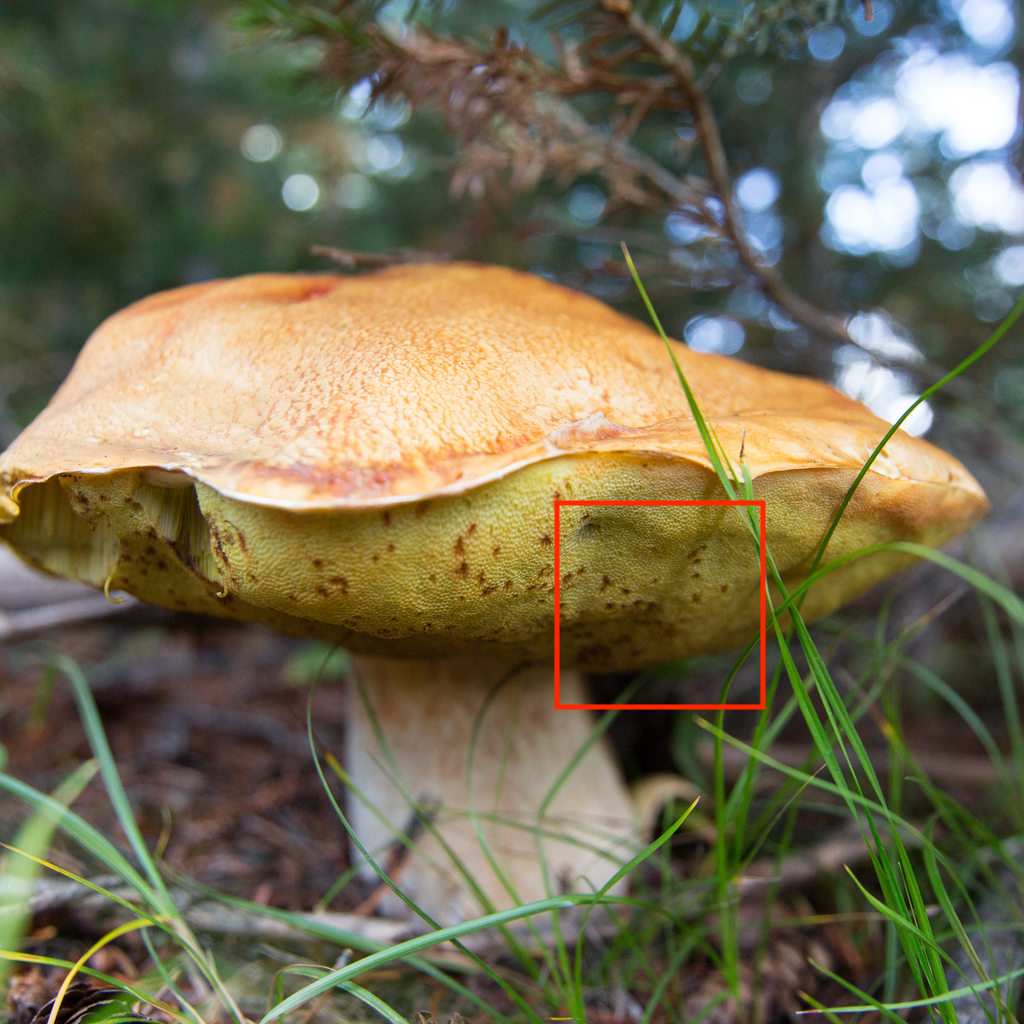} \hspace{-1.mm} &
    \includegraphics[width=0.121\textwidth]{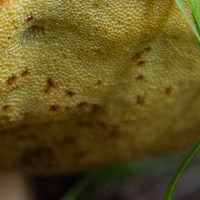} \hspace{-1.mm} &
    \includegraphics[width=0.121\textwidth]{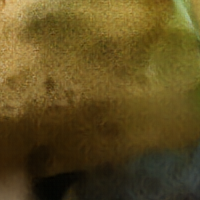} \hspace{-1.mm} &
    \includegraphics[width=0.121\textwidth]{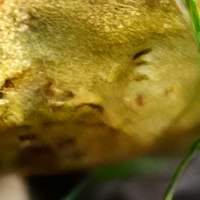} \hspace{-1.mm} &
    \includegraphics[width=0.121\textwidth]{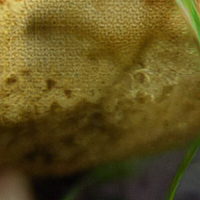} \hspace{-1.mm} &
    \includegraphics[width=0.121\textwidth]{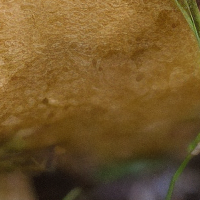} \hspace{-1.mm} &
    \includegraphics[width=0.121\textwidth]{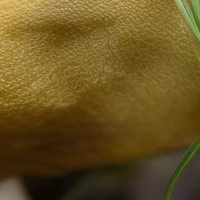} \hspace{-1.mm} &
    \includegraphics[width=0.121\textwidth]{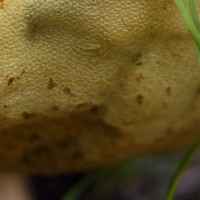} \\
    DIV2K-Val & bpp $\downarrow$ & 0.0420 & 0.0505 & 0.0432 & 0.0535 & 0.0457 & \textbf{0.0322} \\
    \vspace{-2.5mm}
    
    \includegraphics[width=0.121\textwidth]{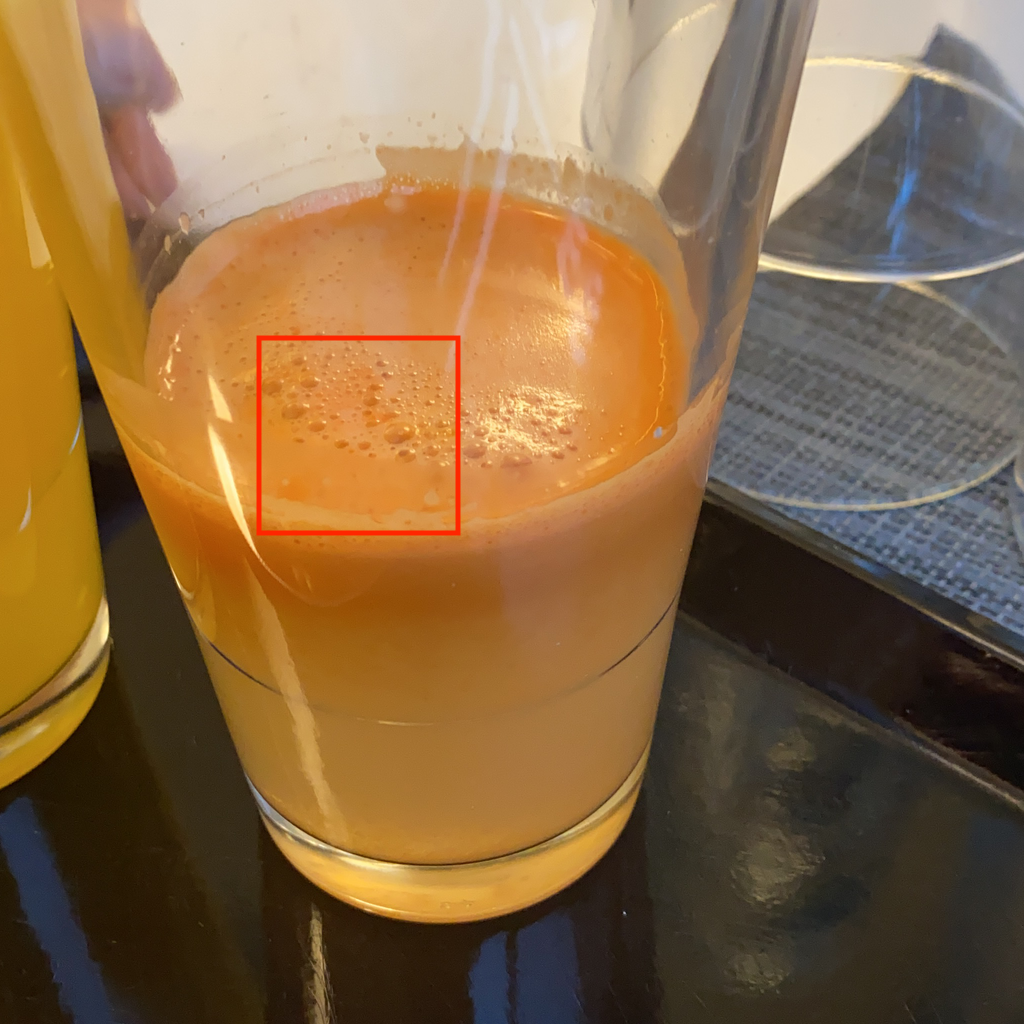} \hspace{-1.mm} &
    \includegraphics[width=0.121\textwidth]{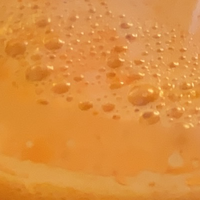} \hspace{-1.mm} &
    \includegraphics[width=0.121\textwidth]{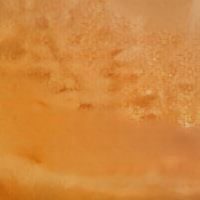} \hspace{-1.mm} &
    \includegraphics[width=0.121\textwidth]{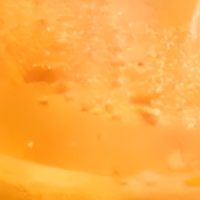} \hspace{-1.mm} &
    \includegraphics[width=0.121\textwidth]{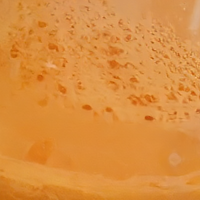} \hspace{-1.mm} &
    \includegraphics[width=0.121\textwidth]{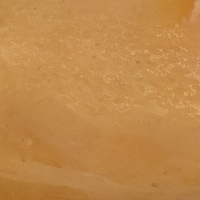} \hspace{-1.mm} &
    \includegraphics[width=0.121\textwidth]{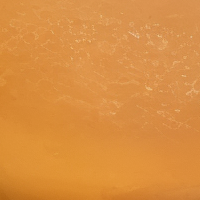} \hspace{-1.mm} &
    \includegraphics[width=0.121\textwidth]{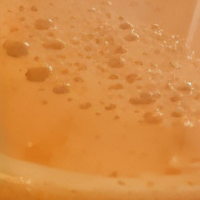} \\
    CLIC2020 & bpp $\downarrow$ & 0.0400 & 0.0472 & 0.0421 & 0.0556 & 0.0462 & \textbf{0.0380}

\end{tabular}
\vspace{-3.mm}
\caption{Qualitative comparison with state-of-the-art methods on the Kodak, DIV2K-Val, and CLIC2020 datasets.}
\label{fig:qualitative_comparison}
\vspace{-3.mm}
\end{figure*}

\paragraph{Stage 3: Joint Training with Rate Annealing.}
In this stage, we perform end-to-end optimization of the entire framework. The pre-trained ViT and the final VAE decoder $\mathcal{D}_m$ remain frozen, while all other networks are trained with full parameters, except the U-Net, which continues to be fine-tuned via LoRA. As the VAE encoder is updated, the latent representation $\hat{y}$ can become distorted. To ensure that the fidelity decoder $\mathcal{D}_a$ continues to produce high-fidelity reconstructions, we introduce an alignment loss, $\mathcal{L}_{align}$. From experiments, we find the MSE loss to be most effective:
\begin{equation}
\label{ackdjfhiuefadsjkcnvmbcdsajkifehgurjb}
    \mathcal{L}_{align} = \mathbb{E}\left[ \| x - \hat{x}_{f} \|^2_2 \right], \quad \text{where} \quad \hat{x}_{f} = \mathcal{D}_a(\hat{y}).
\end{equation}
The experimental details of $\mathcal{L}_{align}$ are provided in the ablation. Then, the training objective becomes:
\begin{equation}
\label{fjodlkcjvhgbncmkdslofijkd}
\mathcal{L}_{\text{overall}} = d(x, \hat{x}) + \lambda \cdot r(\hat{y}, \hat{z}) + \alpha \cdot \mathcal{L}_{align}.
\end{equation}
This objective is to fully leverage the generative power of the diffusion model under the guidance of fidelity-rich features to achieve an optimal rate-distortion-perception trade-off.

Finally, the model is fine-tuned with a GAN-based objective $\mathcal{L}_g$ to enhance the synthesis of rich details while maintaining fidelity. Therefore, the overall loss for this final fine-tuning stage can be written as:
\begin{equation}
\label{yauisodpsfgoijsjnfmkdiuhad}
\mathcal{L}_{\text{finetune}} = d(x, \hat{x}) + \lambda \cdot r(\hat{y}, \hat{z}) + \alpha \cdot \mathcal{L}_{align} + \beta \cdot \mathcal{L}_{g},
\end{equation}
where the hyperparameter $\beta$ is used to control the penalty of the GAN loss. Detailed training hyperparameter settings are provided in the implementation details of the main paper and the supplementary material.

\section{Experiments}
\subsection{Experimental Settings}
\paragraph{Datasets.} 
Our SODEC model is trained using random 512$\times$512 patches extracted from the LSDIR dataset. To evaluate performance, we benchmark SODEC on three standard datasets: Kodak~\cite{kodak_dataset}, DIV2K Validation dataset (denoted as DIV2K-Val), and CLIC2020 test set~\cite{toderici2020clic}. 
We center-crop all images in the validation datasets to 512$\times$512 resolution to facilitate a consistent and fair comparison.

\paragraph{Metrics.}
 Finally, the compression rate is measured in bits per pixel (bpp).
For reconstruction fidelity, we report the PSNR and the MS-SSIM~\cite{wang2003multiscale}. To measure perceptual similarity to the ground truth, we employ LPIPS~\cite{zhang2018unreasonable} and DISTS~\cite{ding2020image}. Furthermore, to evaluate the realism of the generated images in a reference-free setting, we adopt the no-reference metrics NIQE~\cite{mittal2012making} and CLIPIQA~\cite{wang2023exploring}. The compression rate is measured in bits per pixel (bpp).

\paragraph{Implementation Details.}
We choose the HiFiC model~\cite{mentzer2020high} without a discriminator as the VAE compression module. We utilize Stable Diffusion 2.1~\cite{rombach2022high} and set the timestep $t$ as 999 to perform one-step diffusion. We set the batch size to 2 and use the AdamW optimizer with $\beta_1$$=$$0.9$ and $\beta_2$$=$$0.999$. We conduct our experiments on 2 NVIDIA RTX A6000 GPUs. More settings are provided in the supplementary material.

\subsection{Main Results}
We conduct extensive experiments to validate the effectiveness of our one-step diffusion model, SODEC, in the ultra-low bitrate regime. To provide a comprehensive analysis, we benchmark our method against several state-of-the-art generative compression models, covering dominant VAE-based, generative tokenizer paradigms, and multi-step diffusion. Specifically, we compare against MS-ILLM~\cite{muckley2023improving} and HiFiC~\cite{mentzer2020high}, which are leading VAE-based methods. For multi-step diffusion approaches, we compare with CDC~\cite{yang2023lossy}. We also include the current diffusion-based models: PerCo~\cite{korber2024perco} and DiffEIC~\cite{li2024towards}. 

\paragraph{Quantitative Evaluation.}
As shown in the visualized results in Fig.~\ref{fig:main_result}, our proposed SODEC establishes a new state-of-the-art across all evaluated metrics. Our model achieves superior perceptual quality, outperforming other diffusion-based compression models like PerCo~\cite{korber2024perco} and DiffEIC~\cite{li2024towards}. Moreover, our SODEC also excels in reconstruction fidelity (\eg, MS-SSIM).

\paragraph{Qualitative Evaluation.}
We present visual comparisons on three datasets in Fig.~\ref{fig:qualitative_comparison}. SODEC achieves reconstructions closer to the original images. In contrast, existing methods often suffer from missing details or content inconsistencies under extreme compression. More visual comparisons are provided in the supplementary material.

\paragraph{Inference Efficiency.}
Moreover, we compare the inference time in Tab.~\ref{tab:inference_efficiency}. The runtime is tested on one A6000 CPU with the 512$\times$512 image. Our single-step diffusion model, SODEC, offers a substantial advantage in latency. Compared to the multi-step diffusion-based method, PerCo~\cite{korber2024perco}, our SODEC is 26$\times$ faster.

\begin{table}[t]
\centering
\scriptsize 
\setlength{\tabcolsep}{1.7mm} 
\begin{tabular}{lcccc}
    \toprule
    Model & Total Time (ms) & Enc. Time (ms) $\downarrow$ & Dec. Time (ms) $\downarrow$ & bpp $\downarrow$ \\
    \midrule
    HiFiC & 9.3 & 5.4 & 3.9  & 0.0310 \\
    MS-ILLM & 9.3 & 54.5 & 84.4  & 0.0395 \\
    \midrule
    PerCo   & 6,242.2 & 1,540.0 & 4,702.2  & 0.0313 \\
    DiffEIC & 7,827.5 & 266.4 & 7,561.1  & 0.0391 \\
    \midrule
    SODEC & 232.9 & 5.0 & 227.9 & 0.0314 \\
    \bottomrule
\end{tabular}
\caption{Inference efficiency comparison on the DIV2K-Val dataset. Total, encoding, and decoding times are measured on one A6000 GPU with the 512$\times$512 image.}
\vspace{-2.mm}
\label{tab:inference_efficiency}
\end{table}

\subsection{Ablation Study}
We conduct our ablation study on LSDIR (train) and DIV2K-Val (test). By default, the models are trained for 50K steps in the pre-training process, and 40K steps in the SODEC end-to-end training process for a fair comparison.

\begin{table}[t]
\centering
\scriptsize 
\setlength{\tabcolsep}{2.6mm}
\begin{tabular}{lccc}
    \toprule
    Guidance Strategy & MS-SSIM $\uparrow$ & LPIPS $\downarrow$ & bpp $\downarrow$ \\
    \midrule
    (i) No Guidance & 0.8212 & 0.3625 & 0.0424 \\
    (ii) Text Prompt Guidance & 0.8185 & 0.3631 & 0.0412 \\
    (iii) Hyperprior Guidance & 0.8258 & 0.3527& 0.0385 \\
    \midrule
    (iv) Aux. Fidelity Guidance (ours) & 0.8481 & 0.3351 & 0.0368 \\
    \bottomrule
\end{tabular}
\caption{Ablation on the fidelity guidance module.}
\vspace{-4.mm}
\label{tab:ablation_guidance}
\end{table}

\paragraph{Fidelity Guidance Module.}
We conduct an ablation study to validate the effectiveness of our proposed fidelity guidance module. We compare three settings: \textbf{(i)} no explicit guidance; \textbf{(ii)} text prompt guidance (used by PerCo); \textbf{(iii)} semantic features guidance extracted from the hyperprior (used by DiffEIC); and \textbf{(iv)} our fidelity guidance module.

As shown in Tab.~\ref{tab:ablation_guidance}, the baseline model without guidance has poor performance. While using text prompts (case ii) or guidance from the hyperprior (case iii) yields some gains, its impact on reconstruction fidelity is limited. In contrast, our proposed fidelity guidance module leads to a substantial improvement in reconstruction accuracy. Crucially, this significant gain in fidelity is achieved with almost no degradation in perceptual quality as measured by LPIPS. This demonstrates that our guidance mechanism achieves a superior balance between realism and fidelity.

\begin{table}[t]
\centering
\scriptsize 
\setlength{\tabcolsep}{3.5mm}
\begin{tabular}{lccc}
    \toprule
    Alignment Loss Config. & MS-SSIM $\uparrow$ & LPIPS $\downarrow$ & bpp $\downarrow$ \\
    \midrule
    (i) No Alignment Loss & 0.7490& 0.4210& 0.0203\\
    (ii) MSE + LPIPS & 0.7481& 0.3961& 0.0199\\
    (iii) Merged into Main Loss & 0.7984& 0.4023& 0.0232\\
    \midrule
    (iv) MSE only (ours) & 0.7948& 0.3827& 0.0227\\
    \bottomrule
\end{tabular}
\vspace{-1.mm}
\caption{Ablation on the setting of alignment loss ($\mathcal{L}_{align}$).}
\vspace{-2.mm}
\label{tab:ablation_loss}
\end{table}

\begin{table}[t]
\centering
\scriptsize 
\setlength{\tabcolsep}{2.8mm}
\begin{tabular}{lccc}
    \toprule
    Training Strategy & MS-SSIM $\uparrow$ & LPIPS $\downarrow$ & bpp $\downarrow$ \\
    \midrule
    (i) Frozen VAE Module & 0.8512& 0.3761& 0.0695\\
    (ii) Joint Training (Matched bpp) & 0.8621& 0.3750& 0.0678\\
    (iii) Low-to-High bpp Curriculum & 0.8643& 0.3451& 0.0593\\
    \midrule
    (iv) Rate Annealing (ours) & 0.8951& 0.3113& 0.0604\\
    \bottomrule
\end{tabular}
\vspace{-1.mm}
\caption{Ablation study on different training strategies.}
\vspace{-4.mm}
\label{tab:ablation_strategy}
\end{table}

\paragraph{Alignment Loss.}
To ensure the preliminary reconstruction $\hat{x}_f$ remains high-fidelity even as the latent representation $\hat{y}$ gets distorted during fine-tuning, we introduce an alignment loss $\mathcal{L}_align$ to constrain it. We investigate four distinct formulations for this fidelity-preservation mechanism: \textbf{(i)} no alignment loss, where decoder $\mathcal{D}_a$ receives no direct gradient supervision; \textbf{(ii)} a composite loss of perceptual (LPIPS) and distortion (MSE); \textbf{(iii)} no separate $\mathcal{L}_{align}$ term ($\mathcal{L}_{align}$$=$$0$); and \textbf{(iv)} a distortion-only (MSE) loss.

As summarized in Tab.~\ref{tab:ablation_loss}, our results validate the need for an explicit alignment loss, as its omission (case i) significantly degrades performance. While a composite loss (case ii) provides no significant improvement in fidelity, merging the constraint into the main loss (case iii) enhances fidelity but at the expense of perceptual quality. In contrast, a dedicated, distortion-only alignment loss (case iv) substantially boosts fidelity over the composite loss (case ii) with a negligible impact on perception compared with case (ii).

\paragraph{Rate Annealing Training Strategy.}
To validate the efficacy of our proposed rate annealing training strategy, we conduct a comparative analysis of four distinct training schemes: \textbf{(i)} training with the entire VAE compression module frozen, thereby excluding it from the optimization process; \textbf{(ii)} apply a joint training approach, but we manually tune the Lagrange multiplier $\lambda$ to ensure the final bitrate is close to the original values; \textbf{(iii)} a low-to-high bpp curriculum, where the rate penalty is progressively relaxed; and \textbf{(iv)} our proposed high-to-low bpp Rate Annealing strategy.

The results are presented in Tab.~\ref{tab:ablation_strategy}. It is evident that our rate annealing training strategy significantly outperforms all other training schemes. For a given reconstruction quality, our method achieves an average bitrate saving of over 30\%. Conversely, at an equivalent bitrate, our proposed method provides substantially better reconstruction quality. This demonstrates the effectiveness of our approach, which allows the model to first learn a rich feature representation in a less constrained, high-bitrate regime before distilling it into a more efficient, low-bitrate representation.

\section{Conclusion}
In this paper, we address the challenges of high latency and poor fidelity in existing diffusion-based compression models. We propose SODEC, a novel model that demonstrates the effectiveness of single-step diffusion for image compression. We introduce the fidelity guidance module to improve reconstruction fidelity. The module provides explicit structural guidance through high-fidelity preliminary reconstruction. Furthermore, we introduce the rate annealing training strategy that enables effective optimization at extremely low bitrates. Extensive experiments demonstrate that our SODEC achieves excellent rate-distortion-perception performance. Compared with multi-step diffusion approaches, SODEC offers more than 20$\times$ decoding speedup.

\section*{Acknowledgments}
This work was supported by Shanghai Municipal Science and Technology Major Project (2021SHZDZX0102) and the Fundamental Research Funds for the Central Universities.

\bibliography{aaai2026}

\end{document}